\PassOptionsToPackage{protrusion=true,final}{microtype}
\PassOptionsToPackage{T5,T1}{fontenc}
\PassOptionsToPackage{table,xcdraw}{xcolor}
\documentclass[10pt, a4paper]{article}
\usepackage{lrec-coling2024}

\usepackage{tikz-dependency}
\newcounter{treecount}
\newenvironment{tree}[2]
{%
\stepcounter{treecount}
#1~#2 \vspace{-\baselineskip}\\
\noindent
\begin{adjustbox}{max width=1.03\columnwidth, center}
\begin{dependency}[theme=simple, label style={font=\large}, edge style={thick}]}
{\end{dependency}
\end{adjustbox}
}
\newcommand{\translation}[2]{#1 \hfill\mbox{#2}}
\newcommand{\translationlb}[2]{#1\\\phantom{.}\hfill\mbox{#2}}

\DeclareTextSymbolDefault{\ohorn}{T5}
\DeclareTextSymbolDefault{\uhorn}{T5}

\usepackage{tipa}

\usepackage{fontawesome5}
\newcommand*{\fiction}{\faBook}
\newcommand*{\nonfiction}{\faInfoCircle}
\newcommand*{\wiki}{\faWikipediaW}

\newcommand*{\grammarexamples}{\faMagic}

\newcommand*{\social}{\faRss} %

\definecolor{lightblue}{RGB}{206, 232, 240}
\sethlcolor{lightblue}

\usepackage{etoolbox}
\newcommand{\mycircle}[1]{\textcolor{#1}{\faCircle}}
\usepackage{contour}
\contourlength{0.5pt}
\newrobustcmd{\contourcircle}[2]{\raisebox{0.2pt}{\scalebox{0.91}{\contour{#2}{\textcolor{#1}{\faCircle}}}}}

\definecolor{mydarkblue}{RGB}{15, 63, 159}
\definecolor{purple}{RGB}{146, 43, 155}
\definecolor{fuchsia}{RGB}{222, 68, 106}
\definecolor{orange}{RGB}{236, 150, 89}
\definecolor{yellow}{RGB}{236, 229, 85}
\definecolor{darkyellow}{RGB}{196, 173, 57}
\newcommand{\north}{\mycircle{mydarkblue}}
\newcommand{\northcentral}{\mycircle{purple}}
\newcommand{\central}{\mycircle{fuchsia}}
\newcommand{\southcentral}{\mycircle{orange}}
\newcommand{\south}{\contourcircle{yellow}{darkyellow}}
\newcommand{\none}{\mycircle{white}}
\newcommand{\unk}{\textcolor{gray}{\faQuestionCircle}}

\usepackage{booktabs}
\usepackage{adjustbox}

\newlength{\spacer}
\newcommand{\perc}{\kern1pt\%}
\newcommand{\f}{F\textsubscript{1}}

\newcommand{\pos}[1]{\textsc{\MakeLowercase{#1}}}
\newcommand{\rel}[1]{\textit{#1}}
\newcommand{\wrongpos}[1]{\textcolor{red}{\textsc{#1}}}
\newcommand{\gloss}[1]{#1}

\newcommand{\xlmr}{\mbox{XLM-R}}

\newcommand*{\tag}[2]{\textit{#1}\textsubscript{\normalsize\textsc{{\MakeLowercase{#2}}}}}

\newcommand*{\sent}[2] %
{\textit{#2} (#1)}

\definecolor{lightteal}{RGB}{145, 219, 219}
\definecolor{lightyellow}{RGB}{240, 240, 137}

\newcommand{\projname}{MaiBaam}

\newcommand{\lmu}{\faMountain}
\newcommand{\mcml}{\faRobot}
\newcommand{\itu}{\faCompass}

\newcommand{\repourl}{\href{https://github.com/UniversalDependencies/UD_Bavarian-MaiBaam}{\texttt{github.com/\allowbreak{}Universal\allowbreak{}Dependencies/\allowbreak{}UD\_\allowbreak{}Bavarian-\allowbreak{}MaiBaam}}}
\newcommand{\repoissues}{\href{https://github.com/UniversalDependencies/UD_Bavarian-MaiBaam/issues}{\texttt{github.com/\allowbreak{}Universal\allowbreak{}Dependencies/\allowbreak{}UD\_\allowbreak{}Bavarian-\allowbreak{}MaiBaam/\allowbreak{}issues}}}

\title{\projname: \\A Multi-Dialectal Bavarian Universal Dependency Treebank}

\name{Verena Blaschke,\kern-2pt\textsuperscript{\lmu\kern1pt\mcml}
Barbara Kova\v{c}i\'{c},\kern-2pt\textsuperscript{\lmu}
Siyao Peng,\kern-2pt\textsuperscript{\lmu\kern1pt\mcml}\\
\large\bf %
Hinrich Schütze,\kern-2pt\textsuperscript{\lmu\kern1pt\mcml}
Barbara Plank\textsuperscript{\lmu\kern1pt\mcml\kern1.5pt\itu}}

\address{\textsuperscript{\lmu} Center for Information and Language Processing, LMU Munich, Germany \\
  \textsuperscript{\mcml} Munich Center for Machine Learning (MCML), Munich, Germany \\
  \textsuperscript{\itu} Department of Computer Science, IT University of Copenhagen, Denmark  \\
{\tt \{verena.blaschke, b.plank\}@lmu.de}}

\abstract{ %
Despite the success of the Universal Dependencies (UD) project exemplified by its impressive language breadth,  there is still a lack in `within-language breadth': most treebanks focus on standard languages. Even for German, the language with the most annotations in UD,
so far no treebank exists for one of its language varieties spoken by over 10M people:
Bavarian.
To contribute to closing this gap, we present the first multi-dialect Bavarian treebank (\projname) manually annotated with part-of-speech and syntactic dependency information in UD, %
covering multiple text genres (wiki, fiction, grammar examples, social, non-fiction). 
We highlight the morphosyntactic differences between the closely-related Bavarian and German and showcase the rich variability of speakers' orthographies.
Our corpus includes 15k~tokens, covering dialects from all Bavarian-speaking areas spanning three countries. 
We provide baseline parsing and POS tagging results, which are lower than results obtained on German and vary substantially between different graph-based parsers.
To support further research on Bavarian syntax, we make our dataset, language-specific guidelines and code publicly available.
\\\newline}

\begin{document}

\maketitleabstract

\section{Introduction}
In the recent decade, the Universal Dependencies (UD) project (\citealplanguageresource{zeman2023ud-2-13}; \citealp{demarneffe2021ud})  has significantly pushed the frontier in multilingual Natural Language Processing (NLP). UD aims to use consistent syntactic representations for the world's languages. %
As of today, UD provides over 240 treebanks in 140+ languages. Despite this coverage, there is still a gap in `within-language breadth' -- namely, a lack of diversity within high-resource languages and their closely related non-standard languages and dialects. 
For example, while Standard German currently has the largest treebank support in UD (with close to 3.8M annotated words as of UD version~2.13), so far UD lacks a treebank for one of the German language varieties spoken by over 10M people\footnote{%
The exact speaker population is not known, but \citet{rowley2011bavarian} estimates around 11M.
} 
in three different countries: Bavarian. In this paper,  we present \projname,\footnote{EN: `maypole' (lit.\ `May tree'), `Maibaum' in German.} the first UD treebank for Bavarian.

\begin{figure}
    \centering
    \includegraphics[width=\columnwidth, trim={3mm 0 0 3mm}, clip]{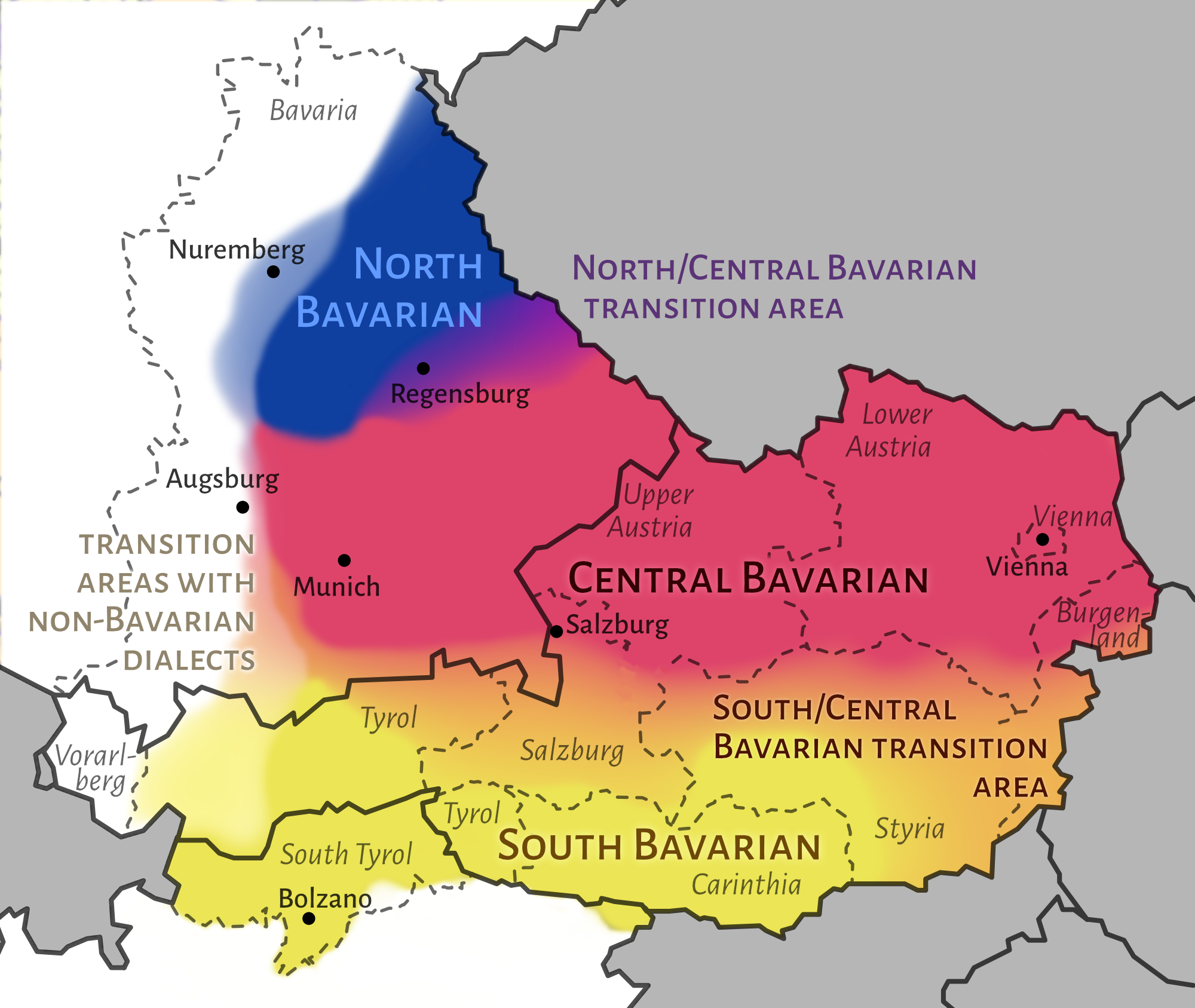}
    \caption{\textbf{Bavarian dialect groups in Germany, Austria and Italy,} based on the classification by \citet[map~47.4]{wiesinger1983einteilung}.
    Names of dialect groups are in \textsc{small caps}, names of provinces and states in \textit{italics.}}
    \label{fig:map}
\end{figure}
Overall, manually annotated corpora are scarce for regional dialects.
In large parts, this is due to the fact that collecting and annotating data for non-standard languages and dialects is especially difficult: it it hard to obtain and collect texts, it is challenging to recruit native-speaking annotators with sufficient linguistic background \citep{miletic-etal-2020-building}, and it requires more expert knowledge and time to adopt guidelines developed for standard languages. %
Despite all of these challenges, collecting and annotating non-standard and dialectal data is rewarding for at least two reasons. From a linguistic perspective, it allows contrastive analyses of language varieties and studying of morphosyntactic differences to standard language. From an NLP perspective, dialects provide a unique test bed with a dire need for technological innovations in light of data scarcity~\cite{blaschke-etal-2023-survey}.

Bavarian is a Germanic language variety closely related to German and spoken in parts of Germany, Austria, and Italy.
The language status of Bavarian is complicated, as it could be called a language distinct from German on linguistic grounds, but is perceived as a dialect by its speakers \cite{rowley2011bavarian}.
It nevertheless has an ISO 639-3 code: \textsc{bar}.
Bavarian comprises a number of local varieties belonging to three major dialect groups: {\north}~North, {\central}~Central, and {\south}~South Bavarian, connected by transition areas to the {\northcentral}~North and {\southcentral}~South of the Central Bavarian area \citep[p.~839]{wiesinger1983einteilung}. %
None of these varieties are standardized. %
Figure~\ref{fig:map} shows where they are spoken and Table~\ref{fig:variation-example} provides an example of the linguistic and orthographic variation.

Our contributions are as follows:
\begin{itemize}
    \item We introduce \projname, the first Bavarian UD treebank (\S\ref{sec:corpus-data}).%
    \item We present annotation guidelines for Bavarian morphosyntactic structures that differ from German ones (\S\ref{sec:tokenization}--\ref{sec:pos-dep}).
    \item We analyse transfer performance of multiple parsers trained on German data (\S\ref{sec:experiments}).
\end{itemize}

\begin{table}
\noindent
\begin{itemize}
  \setlength\itemsep{0em}
    \item[\north] Trotzdean das'e's moch, hairon tou'e's niat.
    \item[\central] Obwoi i's mog, heirodn dua e's ned.
    \item[\south] Trotz dass i's mog, hairatn tua i's net.
    \item[\textsc{deu}] Obwohl ich sie mag, heiraten tu ich sie nicht.
\end{itemize}
\vspace{-\baselineskip}
    \caption{\textbf{Dialectal and orthographic variation in Bavarian.} A \north~North Bavarian grammar example from our corpus, `Although I like her I won't marry her,' translated into \central~Central and \south~South Bavarian as well as German (\textsc{deu}).}
    \label{fig:variation-example}
\end{table}

\noindent
We make our data available at \repourl, to be released in UD~v2.14 (May 2024).
Additionally, we share our guidelines \citep{maibaam-guidelines} and the code we use for preprocessing and annotating the data and for the transfer experiments.\footnote{\url{https://github.com/mainlp/maibaam-code}}

\section{Related Work}
\label{sec:related-work}

The UD project hosts four German treebanks: HDT \citep{borges-volker-etal-2019-hdt}, GSD \citep{mcdonald-etal-2013-universal}, PUD \citep{zeman-etal-2017-conll}, and LIT \citep{salomoni2017lit}.
There also exist two dependency treebanks for related non-standard varieties: the Swiss German UZH \citep{aepli2018parsing} and the Low Saxon LSDC \citep{siewert-etal-2021-towards}, as well as a non-UD Swiss German corpus with phrase structure annotations \citep{schoenenberger2019geparstes}.

A few NLP datasets include Bavarian data.
Both Kontatto \citep{dalnegro2020kontatto} and DiDi \citep{frey2015didi} contain part-of-speech (POS) tags for Bavarian data from South Tyrol; the former was tagged manually or semi-automatically, the latter automatically based on German glosses.
The BarNER dataset provides named entity annotations for Bavarian wiki and social media data \citep{peng2024barner}.
\Citet{van-der-goot-etal-2021-masked} and \citet{winkler2024slot} have collected South and Central Bavarian slot and intent detection data.
The Kontatti corpus \cite{ghilardi2019eliciting} and the multidialectal Zwirner corpus \citeplanguageresource{zwirner} contain Bavarian speech data.

\section{The MaiBaam Treebank}
\label{sec:corpus-creation}

\begin{table}
\begin{adjustbox}{max width=\columnwidth, center}
\begin{tabular}{c@{\hspace{6pt}}r@{\hspace{6pt}}r@{\hspace{6pt}}rc}
\toprule
\textbf{Genre} & \textbf{Tokens} & \textbf{Sents} & \textbf{Tok/Sent} & \textbf{Dialect}\\
\midrule
\wiki & 7\,988 & 417 & 19.2 & \north\northcentral\central\southcentral\south\unk\\
\grammarexamples & 2\,485 & 285 & 8.7 & \north\none\central\phantom{\unk\south}\unk\\
\nonfiction & 2\,019 & 238 & 8.5 & \none\none\central\none\south\unk\\
\social & 1\,599 & 87 & 18.4 & \none\none\none\none\none\unk\\
\fiction & 932 & 43 & 21.7 & \none\none\none\none\none\unk\\
\midrule
All & 15\,023 & 1\,070 & 14.0 & \north\northcentral\central\southcentral\south\unk\\
\bottomrule
\end{tabular}
\end{adjustbox}
\caption{\textbf{Genre distribution in MaiBaam.}
Genres:
\wiki~\textit{Wiki,}
\grammarexamples~\textit{grammar examples,}
\nonfiction~\textit{non-fiction,}
\social~\textit{social,}
\fiction~\textit{fiction.}
Dialects:
\north~\textit{North Bavarian,}
\northcentral~\textit{North/Central B.\ transition dialect,}
\central~\textit{Central B.,}
\southcentral~\textit{South/Central B.\ transition dialect,}
\south~\textit{South B.,}
\unk~\textit{un(der)specified dialect}. %
}
\label{tab:genres}
\end{table}

\subsection{Data and Corpus Statistics}
\label{sec:corpus-data}

Our data come from several different sources that all allow public re-sharing.
They span several UD genres (Table~\ref{tab:genres}):

\begin{itemize}
    \item[\wiki] \textit{Wiki} sentences are taken from Bavarian Wikipedia articles.\footnote{\url{https://bar.wikipedia.org}; CC BY-SA~4.0} 
    We select articles on various different topics to avoid over-representing the template structures that many location-related articles tend to follow.
    Additionally, we nearly exclusively choose articles tagged as being written in a specific dialect.%
    \footnote{\url{https://bar.wikipedia.org/wiki/Kategorie:Artikel_nach_Dialekt}}
    As such, while there still might be some overlapping authors/editors, we expect that the overall set of wiki writers in MaiBaam is fairly diverse.
    \item[\grammarexamples] \textit{Grammar examples} come from three sources: i)~Tatoeba\footnote{\url{https://tatoeba.org/en/sentences/show_all_in/bar/none}; CC BY~2.0 FR} sentences contributed by users who self-report as Bavarian native speakers, ii)~Wikipedia articles that contain collections of linguistic samples, and iii)~UD's Cairo \makebox{CICLing} Corpus,\footnote{\href{https://github.com/UniversalDependencies/cairo}{\texttt{github.com/UniversalDependencies/cairo}}} translated by a Bavarian native speaker.
    \item[\fiction]\textit{Fiction:} We include parts of non-encyclopedic Wikipedia pages recounting fairy tales.
    \item[\nonfiction] The \textit{non-fiction} genre includes questions and commands for a hypothetical digital assistant from the South Tyrolean validation split of xSID \citep{van-der-goot-etal-2021-masked}, as well as from the natural (untranslated) queries and the Central Bavarian test split of xSID from \citet{winkler2024slot}.
    \item[\social] \textit{Social:} We annotate sections of Wikipedia discussion pages and replace usernames mentioned in the text with \textsc{username}.
\end{itemize}

\noindent
Inspired by the analysis of \citet{muller-eberstein-etal-2021-universal}, we annotate each sentence with genre metadata so that relevant patterns can also be analyzed on a genre level rather than only a treebank level.
Table~\ref{tab:genres} shows the distribution of genres in our dataset.
Currently, wiki articles represent the largest group, making up 53\perc{} of the tokens and 39\perc{} of the sentences in MaiBaam.
The average sentence length differs across genres, with fiction, wiki articles and discussions having much longer sentences than grammar examples or queries for digital assistants.
This is consistent with statistics for other treebanks \citep[p.~63]{peng2023cross}.

All of our data sources have metadata indicating that the text is in Bavarian. In many cases, the metadata also mention a more specific dialect or location. 
Table~\ref{tab:dialects} presents the token-level geographical distribution of {\projname} across the dialect groups displayed in Figure \ref{fig:map}.
Just under half of the tokens are in sentences that we can clearly assign to one of the dialect areas.
The {\central}~Central Bavarian group is the dialect area with the most tokens in MaiBaam (22\perc).
This group also contains the two best-represented sub-regions: the cities of Vienna and Munich.

A significant part of our data does not contain any location or dialect information, or refers to larger regions in which multiple dialects are spoken.
In our treebank, we tag each sentence with the most specific dialect and location information available.

We include a full data statement \citep{bender-friedman-2018-data} in Appendix~\ref{sec:data-statement}.
Appendix~\ref{sec:label-distributions} provides an overview of the POS tag and dependency label distributions in our data.
\begin{table}
\begin{adjustbox}{max width=\columnwidth, center}
\begin{tabular}{l@{\hspace{3pt}}l@{\hspace{-6pt}}rr}
\toprule
\multicolumn{2}{l}{\textbf{Dialect group with location}} & \textbf{Tokens} & \textbf{Sents} \\ \midrule
\multicolumn{2}{l}{{\north} \textbf{North Bavarian}} & \textbf{833} & \textbf{65}  \\
\hspace{1em} & Western North Bavarian area & 308 & 34\\
& Unspecified North Bavarian& 525 & 31\\
[2mm]\multicolumn{2}{l}{{\northcentral} \textbf{North/Central Bavarian}} &  \textbf{793} & \textbf{47}\\
 & Bavarian Forest &  793 & 47\\
[2mm]\multicolumn{2}{l}{{\central} \textbf{Central Bavarian}} & \textbf{3\,303} & \textbf{221} \\
 & Munich &  1\,166 & 60\\
 & Cent. Bav. in Upper Bavaria & 613 & 76\\
 & Salzburg (city) & 102 & 5 \\
 & Upper Austria &  43 & 5\\
 & Vienna & 1\,356 & 73 \\
 & Unspecified Central Bavarian& 23 & 2 \\
[2mm]\multicolumn{2}{l}{\textbf{{\southcentral} South/Central Bavarian}} &  \textbf{1\,130} & \textbf{50} \\
& Bad Reichenhall & 206 & 11\\
 & Berchtesgaden & 110 & 5\\
 & Pongau &  515 & 21\\
 & Pinzgau &  299 & 13\\
[2mm]\multicolumn{2}{l}{{\south} \textbf{South Bavarian}} & \textbf{995} & \textbf{70} \\
 & Carinthia & 99 & 4 \\
 & South Tyrol & 896 & 66\\
\midrule
\multicolumn{2}{l}{\unk{} \textbf{Underspecified}} & \textbf{7\,969} & \textbf{617} \\
& {\central\southcentral\south?} Upper Bavaria & 438 & 28\\
& {\central\southcentral\south?} Austria & 1\,491 & 173\\
& {\central\southcentral\south?} Other C., S./C. or S. & 901 & 122\\
& {\central\southcentral?} South East Upper B. & 182 & 10\\
& {\central\southcentral?} East Austria & 1\,687 & 86\\
& {\southcentral\south?} Styria &  292 & 15\\
& {\north\northcentral\central\southcentral\south?} Unspecified & 2\,980 & 183\\
\bottomrule
\end{tabular}
\end{adjustbox}
    \caption{\textbf{Dialect groups in MaiBaam.}
    `Underspecified' refers to cases where we do not have any dialect or location information (`unspecified') or where the specified geographic area encompasses multiple dialect groups.}
    \label{tab:dialects}
\end{table}

\subsection{Annotation Procedure}
\label{sec:annotation-procedure}
The annotation procedure 
includes training and adjudication sessions, first on a sample of German texts from existing UD treebanks~(\S\ref{sec:related-work}) and then on the target Bavarian data.
We train the annotator initially on universal part-of-speech (UPOS) tags and later also on dependencies.
We use a modified version of ConlluEditor \citep{heinecke-2019-conllueditor} to annotate the data.

The annotator is a computational 
linguistics student who is a native speaker of German and a (non-Bavarian) Upper German dialect.
We also consult three Bavarian native speakers from the South and Central Bavarian dialect areas for grammaticality judgments, lexical disambiguation and translations.
The annotator and consultants involved in this project are hired and compensated according to local standards.

Manually correcting automatically predicted labels is a common strategy in UD annotation to save time and labour \citep{salomoni2017lit, borges-volker-etal-2019-hdt}.
To minimize bias from model outputs, we use a simple rule-based pre-tokenizer.
Inspired by \citet{berzak-etal-2016-anchoring}, we pre-annotate POS tags where the UDPipe model trained on GSD and the one trained on HDT agree and jointly achieve a precision >95\perc{} on an initial test set of 4k tokens.\footnote{The POS tags with high precision scores are: \pos{aux, cconj, det, noun, num, part, pron,} and \pos{punct.}
About 41\perc{} of the tokens in the documents we pre-annotate receive a POS tag.} 
We do not pre-annotate dependency arcs or labels.
In terms of dependency labels, the ones that can easily be predicted are also the ones that can trivially be expressed in a rule-based way as suggestions within ConlluEditor (e.g., if a \pos{det} is marked as a dependent of a \pos{noun}, the relation will be \rel{det}).

We largely follow UD's annotation guidelines for  German.\footnote{\url{https://universaldependencies.org/de}}
To the extent it is possible, we also follow the annotation decisions made in the closely related treebanks (\S\ref{sec:related-work}), which however often disagree in more particular grammatical contexts.\footnote{This observation is not new; see \citet{hovy-etal-2014-pos} and \citet{wisniewski-yvon-2019-bad} for investigations of POS inconsistencies (the latter specifically within UD) and \citet{zeldes-schneider-2023-ud} for a comparison of decisions made in two large English UD treebanks. Our annotation guidelines contain more details on the German case.} When the grammatical structures are similar to English ones, we also consult the English EWT \citep{silveira-etal-2014-gold} and GUM \citep{zeldes2017gum} treebanks.
We use Grew-match \citep{guillaume-2021-graph} for querying these treebanks.

We discuss and resolve difficult annotation cases in weekly meetings.
As additional approaches to finding annotation errors, we use Udapi \citep{popel-etal-2017-udapi} and UD's validation scripts.\footnote{\url{https://github.com/UniversalDependencies/tools}}
Furthermore, we manually double-check the word forms that we annotated with closed-class POS tags.

The annotation time -- excluding training time, research into the annotation decisions made for other treebanks, discussions of grammatical phenomena, adjudication meetings and subsequent corrections -- totals 165~hours.
The average annotation time per sentence varies greatly, depending on the text genre and dialect, as well as on the level of familiarity with the guidelines and annotation tools.

We decide against normalizing the data to an artificial Bavarian standard since no actual written, or even spoken, standard exists. 
Such a decision would ultimately have been biased towards certain Bavarian dialects, thus conflicting with our goal of curating a diverse set of Bavarian varieties. 

\subsection{Tokenization}
\label{sec:tokenization}

In Bavarian, prepositions and determiners are often contracted, e.g., \textit{beim} `at the.\textsc{dat}'.
We follow the UD guidelines for German (see also \citealp{grunewald-friedrich-2020-unifying}) and treat such cases as multi-word tokens: \textit{beim} becomes \tag{bei}{adp} plus \tag{m}{det}.
This decision is also consistent with how the Low Saxon guidelines \citep{siewert-etal-2022-low}\footnote{\url{https://universaldependencies.org/nds}} handle tokenization
but differs from the decision made for Swiss German to leave merged word sequences as they are written \citep{aepli2018parsing}.\footnote{\url{https://universaldependencies.org/gsw}}
Since there is variation in the way determiners are pronounced and written, we simply split the words into substrings (rather than normalizing them to an arbitrary standard).\footnote{Full forms of the dative definite determiner in our corpus include \textit{dem, am, im} and, due to partial case syncretism of dative and accusative forms \citep[pp.~85, 98]{merkle1993bairische}, also \textit{den, an, in}.}

Other commonly fused sequences that we split are determiners followed by common or proper nouns (\tag{d'}{det} \tag{Rundn}{noun} `the round') and verbs or complementizers followed by pronouns or neuter determiners (\tag{houd}{aux} \tag{s}{det} `has the'; \tag{habn}{aux} \tag{se}{pron} \tag{s}{pron} `they have [...] it').

When a vowel-initial word is appended to a vowel-final word, a linking consonant can be inserted in between \citep[pp.~30--33]{merkle1993bairische}.
In this case, we include the consonant with the first word (e.g., we analyze \textit{wiera} `how he' with its linking \textit{-r-} as \tag{wier}{sconj} and \tag{a}{pron}).

In order to enable comparisons with datasets tokenized like the Swiss German UD treebank,
we will include a script upon data release that reverts the token splits, assigns tags to the unsplit tokens (e.g., \pos{det+noun} becomes \pos{noun} and \pos{verb+pron} becomes \pos{verb}) and adjusts the dependencies accordingly.

When it comes to hyphenated compound words, we follow the German HDT treebank and do \textit{not} split them apart: e.g., \textit{Fabel-Viech} `mythical creature' is a single word.

\subsection{POS Tags and Dependencies}
\label{sec:pos-dep}

We use the same set of dependency relations and subrelations\footnote{These subrelations include dative objects \rel{(obl:arg)}, possessive pronouns \rel{(det:poss)}, lexicalized reflexive pronouns \rel{(expl:pv)}, particle verbs \rel{(compound:prt)}, passive constructions \rel{(nsubj:pass, csubj:pass, aux:pass, obl:agent)} and relative clauses \rel{(acl:relcl, advcl:relcl).}} as defined for German and refer to the German guidelines and treebanks where possible.\footnote{%
In cases not mentioned by the guidelines and where the German treebanks disagree, we make decisions based on other sources.
For instance, in the case of \textit{selbst/\allowbreak{}selber} `self' being added to a sentence for emphasis, we follow the analysis by \citet{hole-2002-selbst} who distinguishes between \textit{selbst} being used as an adnominal or adverbial intensifier, and attach the word to the corresponding noun or clausal head respectively. 
Other such cases are detailed in \citet{maibaam-guidelines}.
}
However, Bavarian permits syntactic structures that are not licensed in Standard German.\footnote{Analytic possession and articles before person names do appear in colloquial German, but are uncommon in written Standard German.}
We here discuss a range of such structures in our data, along with our annotation decisions.

\subsubsection{Verbs}

\paragraph{Infinitives}
In German, many infinitive constructions require the marker \tag{zu}{part}.
In Bavarian, two similar constructions appear: one where a cliticized form of the marker \textit{(z)} is followed by a verbal infinitive,
and one where the infinitive is nominalized and the a cliticized dative determiner \textit{(m/n)} is added to the marker: \textit{zum} or \textit{zun} \citep{bayer1993zum, bayer2004klitisiertes-zu}.
In both cases, we annotate \tag{z(u)}{part} with \rel{mark} (as in the German treebanks), and in the latter, we separately annotate \tag{m/n}{det} with \rel{det}:

\vspace{0.5\baselineskip}
\stepcounter{treecount}
\noindent
\begin{adjustbox}{minipage=\columnwidth, center} %
(\thetreecount{}) \scalebox{0.9}{Ludwig van Beethoven hod de Gwohnheit ghobt,}\\
\phantom{(1)} \scalebox{0.9}{\gloss{Ludwig van Beethoven had had the habit}}

\noindent
\scalebox{0.9}{
\begin{dependency}[theme=simple, label style={font=\large}, edge style={thick}]
\begin{deptext}
genau 60 Kafääbaunan \& zu\& m \& oozöön\&[-4pt] , \\
\gloss{exactly 60 coffee beans} \& \textsc{inf} \& \gloss{the} \& \gloss{count} \& ,\\
\& \pos{part}\& \pos{det}\& \pos{noun}\&\&\\
\end{deptext}
\depedge[]{4}{2}{mark}
\depedge[label style={below}]{4}{3}{det}
\end{dependency}
}

\vspace{-5pt}
\noindent\hspace{-3mm} 
\scalebox{0.9}{
\begin{dependency}[theme=simple, label style={font=\large}, edge style={thick}]
\begin{deptext}
 um \&[-2pt] si \&[-2pt] draus\&[-2pt]  a Schalal Mokka\& z \& mochn \&[-4pt] . \\
\gloss{so as to} \& \textsc{refl} \& \gloss{out of it} \&  \gloss{a cup of coffee} \& \textsc{inf} \& \gloss{make} \& .\\
\&\&\&\& \pos{part}\& \pos{verb}\\
\end{deptext}
\depedge[right=2mm]{6}{5}{mark}
\end{dependency}
}
\end{adjustbox}
\translation{`Ludwig van Beethoven had a habit of counting exactly 60 coffee beans in order to brew a cup of coffee from them.'}{(Wiki \textit{Kafää} `Coffee')}

\paragraph{Auxiliary \textit{tua} `do'}
In addition to the auxiliary verbs named in the German guidelines, we include \textit{tua} `do', which is used in several periphrastic constructions in conjunction with a lexical verb, both in indicative and subjunctive constructions \citep[pp.~65--67]{merkle1993bairische}.

\begin{tree}{(\thetreecount{})}{}
\begin{deptext}
Waun i du wa, \& tarat \& i \& 'n \& frogn \&.\\
\gloss{If I were you,} \& \gloss{do}.\textsc{1sg.sbjv} \& \gloss{I} \& \gloss{him} \& \gloss{ask} \& .\\
\& \pos{aux} \& \pos{pron} \& \pos{pron} \& \pos{verb} \& \\
\end{deptext}
\depedge[label style=below, arc angle=30]{5}{2}{aux}
\end{tree}
\translationlb{`If I were you, I would ask him.'}{(Tatoeba 5166978)}

\subsubsection{Noun Phrases}
\label{sec:np}
\paragraph{Order of determiner and adverb} 
In German, if an adverb modifies an adjective in a noun phrase, the adverb appears between the determiner and the adjective: 

\addtocounter{treecount}{-1}
\begin{tree}{}{}
\begin{deptext}
\pos{det} \&[3pt] \pos{adv} \&[3pt] \pos{adj} \&[3pt] \pos{noun} \\
\end{deptext}
\depedge[edge start x offset=6pt]{4}{1}{det}
\depedge[left=3mm]{3}{2}{advmod}
\depedge[edge start x offset=6pt, left=2mm]{4}{3}{amod}
\end{tree}

\vspace{-\baselineskip}
\noindent
For a small set of Bavarian intensifiers, alternative orders are possible (typically when the determiner is indefinite): the order of adverb and determiner can be reversed (\pos{adv~det~adj~noun}) and the determiner can be doubled (\pos{det~adv~det~adj~noun}) (\citealp{lenz2014dynamik}; \citealp[pp.~89--90, 158]{merkle1993bairische}).
In such cases, we allow non-projective dependencies:

\begin{tree}{(\thetreecount)}{}
\begin{deptext}
Frier wor des \& gonz \&a \&normales \&Wort \\
\gloss{Previously, it was} \& \gloss{very} \& \gloss{a} \& \gloss{normal} \& \gloss{word}\\
\& \pos{adv} \& \pos{det} \& \pos{adj} \& \pos{noun} \\
\end{deptext}
\depedge{5}{3}{det}
\depedge[left=2mm]{4}{2}{advmod}
\depedge[edge style={gray}, label style={text=gray,below}]{5}{4}{amod}
\end{tree}
\translationlb{`It used to be a completely normal word.'}{(Wiki  \textit{Walsch} `Italian/Romance')}

\vspace{0.5\baselineskip}
\begin{tree}{(\thetreecount)}{}
\begin{deptext}
In da englischn is \&[-6pt] a \& ganz \& a \& bläds \& Buidl \& drin \\
\gloss{In the English one is} \& \gloss{a} \& \gloss{very} \& \gloss{a} \& \gloss{silly} \& \gloss{picture} \& \gloss{inside} \\
\& \pos{det} \& \pos{adv} \& \pos{det} \& \pos{adj} \& \pos{noun} \\
\end{deptext}
\depedge{6}{2}{det}
\depedge{6}{4}{det}
\depedge{5}{3}{advmod}
\depedge[edge style={gray}, label style={text=gray,below}]{6}{5}{amod}
\end{tree}
\translationlb{`The English [wiki] contains a very silly picture [...]'}{(Wiki discussion \textit{Ottoman} `sofa')}

\paragraph{Postponed adjectives}
For emphasis (and especially when voicing annoyance), phrases of the pattern \pos{(adp) det adj noun} can be rearranged into \pos{(adp) det noun (adp) det adj} \citep[p.~168]{merkle1993bairische}.
We consider the postponed adjective to be an apposition of the noun.
This structure is often combined with constructions where a first or second-person pronoun is used in lieu of a determiner.
In such cases, we tag the pronoun as \pos{pron} and, following the recommendation by \citet{hohn-2021-towards}, label the relation \rel{det}.
In the following sentence in our corpus (pardon our Bavarian), \textit{du bleda Depp} `you stupid idiot' is re-arranged:

\begin{tree}{(\thetreecount)}{}
\begin{deptext}
Hau di üba d'Heisa, \& du \& Depp  \& du \& bleda \& ! \\
\gloss{Get lost,}\& \gloss{you} \& \gloss{idiot} \& \gloss{you} \& \gloss{stupid} \& \gloss{!}\\
\& \pos{pron} \& \pos{noun} \& \pos{pron} \& \pos{adj} \\
\end{deptext}
\depedge{3}{2}{det}
\depedge[arc angle=90]{3}{5}{appos}
\depedge[left=2mm]{5}{4}{det}
\end{tree}
\translation{`Get lost [lit.: scram over the houses], you stupid idiot!'}{(Tatoeba 5657152)}

\paragraph{Personal names} In Bavarian, personal names are preceded by a determiner matching in case and gender \citep[pp.~69--70]{weiss1998syntax}, and the family name is often put before the given name \citep[p.~71]{weiss1998syntax}.
Following the general UD guidelines, we connect the parts of the name via a \textit{flat} relation:

\begin{tree}{(\thetreecount)}{}
\begin{deptext}
weder \& da\&  Schmidt\&  Bäda \& no \& d’\& Braun \& Maria\\
\gloss{neither} \& \gloss{the}.\textsc{m} \& \gloss{Smith} \& \gloss{Peter} \& \gloss{nor} \& \gloss{the}.\textsc{f} \& Brown \& Mary\\
\pos{cconj} \& \pos{det} \& \pos{propn} \& \pos{propn} \& \pos{cconj} \& \pos{det} \& \pos{propn} \& \pos{propn} \\
\end{deptext}
\depedge[label style={font=\Large}]{3}{2}{det}
\depedge[label style={font=\Large}]{3}{4}{flat}
\depedge[label style={font=\Large}]{7}{6}{det}
\depedge[label style={font=\Large}]{7}{8}{flat}
\end{tree}
\translationlb{`neither Peter Smith nor Mary Brown [...]'}{(Cairo CICLing~12)}

\paragraph{Possession} Bavarian, like many German dialects and colloquial variants, eschews the genitive in favour of analytic possessive constructions \citep{fleischer2019syntax, buelow2021structures}.
These can be prepositional phrases or the prenominal dative construction, in which we analyze the possessor as an \textit{nmod}:

\begin{tree}{(\thetreecount{})}{\vspace{-\baselineskip}}
\begin{deptext}
ohn \& in \& Lutha \& seina \& Iwasezung \\
\gloss{without} \& \gloss{the}.\textsc{dat} \& \gloss{Luther} \& \gloss{his} \& \gloss{translation} \\
\pos{adp} \& \pos{det} \& \pos{propn}\& \pos{det}\& \pos{noun}\\
\end{deptext}
\depedge{3}{2}{det}
\depedge[edge start x offset=3pt, edge end x offset=-3pt, label style={below}, below=1mm]{5}{4}{det:poss}
\depedge[edge start x offset=3pt]{5}{3}{nmod}
\end{tree}
\translationlb{`[...] without Luther's translation [...]'}{(Wiki discussion \textit{Ödenburg} `Sopron')}

\subsubsection{Subordinate Clauses}

\paragraph{Relative markers}
Where German uses the relative pronouns \textit{der/die/das} `that, which', Bavarian can append the invariant relative marker \textit{wo} (in some dialects \textit{was}) \citep{moser2023relative-particles}.
We tag the relative pronoun as \pos{pron} (as in the German treebanks) and the relative marker as \pos{sconj} with the relation \textit{mark}:

\begin{tree}{(\thetreecount{}a)}{\vspace{-\baselineskip}}
\begin{deptext}
'S gibt owa no vui \&[-8pt] Junge \&[-6pt], \&[-2pt] de \&[-13pt]  wo \&[-8pt]  s' \&[-6pt] Boarische  \&[-3pt] no \&[-4pt] vastenga\\
\gloss{But there are}  \& \gloss{young} \&, \& \textsc{rel}. \& \textsc{rel} \& \gloss{the} \& \gloss{Bavarian} \& \gloss{still} \& \gloss{under-} \\[-3pt]
\gloss{still many}\& \gloss{ones}.\textsc{acc} \& \&  \textsc{3pl.nom} \& \& \& \& \& \gloss{stand}\\[-3pt]
\& \& \& \hspace{-3mm}\pos{pron} \& \pos{sconj} \& \hspace{2mm}\pos{det} \&  \pos{noun} \&  \pos{adv} \&  \pos{verb} \\
\end{deptext}
\depedge[label style={font=\Large,below}, below=2mm, arc angle=60]{9}{5}{mark}
\depedge[label style={font=\Large}, arc angle=65]{9}{4}{nsubj}
\depedge[edge style={gray}, label style={font=\Large, text=gray}]{2}{9}{acl:relcl}
\end{tree}
\translationlb{`However, there are still many young people who still understand Bavarian [...]'}{(Wiki \textit{Minga} `Munich')}

\noindent
In certain situations, the relative pronoun can be dropped in Bavarian if the relative marker is present \citep{pittner1996attraktion}. 
This can for instance happen when the relative pronoun would be in the nominative case: %

\addtocounter{treecount}{-1}
\begin{tree}{(\thetreecount{}b)}{\vspace{-\baselineskip}}
\begin{deptext}
'S gibt owa no vui \&[-8pt] Junge \&[-8pt], \&[-3pt]  wo \&[-2pt]  s' \&[-3pt] Boarische  \&[-1pt] no \&[-2pt] vastenga\\
\gloss{But there are}  \& \gloss{young} \&, \& \textsc{rel} \& \gloss{the} \& \gloss{Bavarian} \& \gloss{still} \& \gloss{understand} \\[-3pt]
\gloss{still many}\& \gloss{ones}.\textsc{acc} \\[-10pt]
\& \& \& \pos{sconj} \& \pos{det} \&  \pos{noun} \&  \pos{adv} \&  \pos{verb} \\
\end{deptext}
\depedge[label style={font=\Large,below}, below=2mm, arc angle=60]{8}{4}{mark}
\depedge[edge style={gray}, label style={font=\Large,text=gray}]{2}{8}{acl:relcl}
\end{tree}

\paragraph{Additional complementizer}
The adverb, relative pronoun, or question word introducing a subordinate clause can be followed by an additional conjunction \textit{dass} `that'
(\citealp[pp.~29--30]{weiss1998syntax}; \citealp[pp.~190--191]{merkle1993bairische}), which we consider a \rel{mark}er:

\begin{tree}{(\thetreecount{})}{\vspace{-\baselineskip}}
\begin{deptext}
Jezz mechad i owa wissn, \&[-3pt] wia \& lang \&[-2pt] das \&[-3pt] des \&[-3pt] no \& dauat \&[-4pt] .\\
\gloss{Now I'd like to know} \& \gloss{how} \& \gloss{long} \& \gloss{that} \& \gloss{this} \& \gloss{still} \& \gloss{takes} \& .\\
\& \pos{adv} \& \pos{adj}\& \pos{sconj}\& \pos{pron}\& \pos{adv}\& \pos{verb} \&\\
\end{deptext}
\depedge[left=2mm]{3}{2}{advmod}
\depedge{7}{3}{advmod}
\depedge[label style=below]{7}{4}{mark}
\end{tree}
\translationlb{`Now I'd like to know how long this will still take.'}{(Wiki \textit{Pronomen} `Pronouns')}

\paragraph{Complementizer agreement}
In Bavarian, reduced forms of 2nd person (and, optionally, \textsc{1pl}) pronouns are used when they appear in the Wackernagel position immediately after complementizers.
These reduced forms are immediately attached to the previous word and can still be followed by a full pronoun for additional stress \citep[p.~119]{weiss1998syntax}.
Whether these constructions should be analyzed as a word followed by an enclitic pronoun or as inflected complementizers is debatable (for an overview of the different arguments, see \citealp[pp.~123--133]{weiss1998syntax}).
For our annotations, we follow \citet{bayer2013klitisierung} and adopt the interpretation of inflection:

\begin{tree}{(\thetreecount{})}{\vspace{-0.5\baselineskip}}
\begin{deptext}
Er wüll, \& das'st \& (Du) \& redst \& .\\
\gloss{He wants} \& \gloss{that}.\textsc{2sg} \& \gloss{you}.\textsc{sg} \& \gloss{talk}.\textsc{2sg} \& .\\
\& \pos{sconj} \& \pos{pron} \& \pos{verb} \& \\
\end{deptext}
\depedge[label style=below]{4}{2}{mark}
\depedge[label style=below]{4}{3}{nsubj}
\end{tree}
\translationlb{`He wants you to talk.'}{(Wiki \textit{Konjunktiona} `Conjunctions')\vspace{-\baselineskip}}

\vspace{\baselineskip}
\begin{tree}{(\thetreecount{})}{}
\begin{deptext}
Er wüll, \& das \& i \& redt \& .\\
\gloss{He wants} \& \gloss{that} \& \gloss{I} \& \gloss{talk}.\textsc{1sg} \& .\\
\& \pos{sconj} \& \pos{pron} \& \pos{verb} \& \\
\end{deptext}
\depedge{4}{2}{mark}
\depedge[label style=below]{4}{3}{nsubj}
\end{tree}
\translationlb{`He wants me to talk.'}{(Wiki \textit{Konjunktiona} `Conjunctions')\vspace{-\baselineskip}}

\subsubsection{Other}

\paragraph{Dropped 2nd person pronouns}
Similarly, second person pronouns can be omitted when they occur after a correspondingly inflected verb.
Consider for instance the sentence \textit{Kaunst du aufstehn?} `Can you get up?' where the pronoun \textit{du} `you.\textsc{sg}' can be dropped:

\begin{tree}{(\thetreecount{})}{}
\begin{deptext}
Kaunst \& aufstehn \& ?\\
\gloss{Can}.\textsc{2sg} \& \gloss{get up}.\textsc{inf} \& ?\\
\pos{aux} \& \pos{verb} \& \\
\end{deptext}
\depedge[edge style={gray}, label style={text=gray,below}]{2}{1}{aux}
\end{tree}
\translation{`Can you get up?'}{(Tatoeba 10673747c)}

\paragraph{\textsc{1pl} inflection endings}

The \textsc{1pl.pres} inflection of verbs is typically straightforward, e.g., \textit{mia schbui+n} `we play+\textsc{1pl}'.
However, it is also possible to add \textit{-ma} to the stem: \textit{mia schbui+ma} \citep[p.~127]{merkle1993bairische}.
Although this ending historically comes from a cliticized form of the pronoun (and some analyze it as such; \citealp[p.~123, fn.~48]{weiss1998syntax}), we simply analyze it as inflection: \tag{mia}{pron} \tag{schbuima}{verb}.
This decision also lends itself well to UD annotation: if we were to treat \textit{-ma} as its own word, it is unclear what dependency label it should get, since the independent pronoun \textit{mia} is already the subject.

\paragraph{Negative concord} 
Unlike German, Bavarian allows for negative concord in constructions with (inflected forms of) \textit{koa} `no' \citep[pp.~167--168]{weiss1998syntax}:

\begin{tree}{(\thetreecount)}{\vspace{-\baselineskip}}
\begin{deptext}
Se \& hom \& koane \& Haxn \& ned \\
\gloss{They} \& \gloss{have} \& \gloss{no} \& \gloss{legs} \& \gloss{not} \\
\pos{pron} \& \pos{verb} \& \pos{det} \& \pos{noun} \& \pos{adv} \\
\end{deptext}
\depedge[edge style={gray}, label style={text=gray}]{2}{1}{nsubj}
\depedge[edge style={gray}, left=2mm, label style={text=gray, below}]{2}{4}{obj}
\depedge{2}{5}{advmod}
\depedge[label style=below]{4}{3}{det}
\end{tree}
\translation{`They have no legs [...]'}{(Wiki \textit{Fiisch} `Fish')}

\section{Transfer Experiments}
\label{sec:experiments}
This section establishes baselines to gauge how well dependency parsers trained on German perform on our Bavarian treebank. 

\subsection{Data}

Only two German UD treebanks (v2.12) include a training partition: HDT, which comprises almost 3.5M words from web news articles, and GSD, which contains 292k words from news articles, reviews, Wikipedia articles and other webpages.
Both are released under a \href{https://creativecommons.org/licenses/by-sa/4.0/}{CC BY-SA} license.
We train and validate on HDT and GSD and test on the entire gold-tokenized \projname.\footnote{Since {\projname} contains various multi-word tokens (\S\ref{sec:tokenization}), this makes for an easier test condition than if we tested systems that include a tokenization step.
Using the UDPipe models on plain text input, the tokenization {\f} scores are 96.76\perc{} for the version trained on GSD, and 95.14\perc{} for the HDT version.}

\subsection{Models}
\label{sec:models}
\begin{table}
\begin{adjustbox}{max width=\columnwidth, center}
\begin{tabular}{@{}l@{\hspace{3pt}}l@{\hspace{-35pt}}r@{}}
\toprule
\textbf{System or}& \textbf{Pretraining}\\
\textbf{language model} & \textbf{language(s)} & \textbf{License} \\
\midrule
\href{https://lindat.mff.cuni.cz/services/udpipe/}{UDPipe} 2.12-230717  & \textit{see} & \href{https://creativecommons.org/licenses/by-nc-sa/4.0/}{CC BY-NC-SA} \\
\hspace{6pt}\citep{straka-2018-udpipe}&\textit{mBERT}\\[4pt]
\href{https://stanfordnlp.github.io/stanza/}{Stanza} 1.6.1 \citep{qi-etal-2020-stanza} & \textsc{deu} & \href{https://www.apache.org/licenses/LICENSE-2.0}{Apache 2.0}\\
\midrule
\multicolumn{2}{@{}l}{\href{https://machamp-nlp.github.io/}{MaChAmp} \citep{van-der-goot-etal-2021-massive}} & \href{https://github.com/machamp-nlp/machamp/blob/master/LICENSE}{Apache 2.0} \\
\multicolumn{2}{@{\hspace{6pt}}l}{\textit{with the language models below:}} \\[4pt]
mBERT \citep{devlin-etal-2019-bert} & multi, incl. & \href{https://www.apache.org/licenses/LICENSE-2.0}{Apache 2.0} \\
\hspace{6pt}\href{https://huggingface.co/bert-base-multilingual-cased}{bert-base-multilingual-cased} & \textsc{deu} \& \textsc{bar}\\[4pt]
XLM-R \citep{conneau-etal-2020-unsupervised} & multi, incl. \textsc{deu} & \href{https://opensource.org/license/mit/}{MIT}\\
\hspace{6pt}\href{https://huggingface.co/xlm-roberta-base}{xlm-roberta-base}\\[3pt]
GBERT \citep{chan-etal-2020-germans} & \textsc{deu} & \href{https://opensource.org/license/mit/}{MIT}\\
\hspace{6pt}\href{https://huggingface.co/deepset/gbert-base}{deepset/gbert-base}\\
\bottomrule
\end{tabular}
\end{adjustbox}
\caption{\textbf{Systems and pretrained language models used for parsing experiments.}
Key: {\textsc{deu}} is German, {\textsc{bar}} is Bavarian. All systems are finetuned on German data.}
\label{tab:models}
\end{table}

We compare the POS tagging and dependency parsing results of several models.
Version and license details can be found in Table~\ref{tab:models}.

This includes two already trained systems, each trained once on GSD and once on HDT. 
Firstly, we use {UDPipe} \citep{straka-2018-udpipe}, which combines mBERT embeddings \citep{devlin-etal-2019-bert}\footnote{\url{https://ufal.mff.cuni.cz/udpipe/2/models\#universal_dependencies_212_models}}
with custom word and character embeddings.
The architecture of UDPipe is specifically built for dependency parsing and contains steps for ensuring a proper graph structure of the predicted dependencies.
Secondly, we investigate the predictions made by {Stanza} \citep{qi-etal-2020-stanza}, which uses German word embeddings \citep[via][]{zeman-etal-2017-conll} and includes a graph-based dependency parser.

For comparison, we use MaChAmp \citep{van-der-goot-etal-2021-massive}
 to train models from scratch. %
We finetune the multilingual models mBERT and {\xlmr} \citep{conneau-etal-2020-unsupervised} as well as the German model GBERT \citep{chan-etal-2020-germans}, and otherwise use MaChAmp's default settings.
Both multilingual models contain German data in their pretraining data, and mBERT was also pretrained on Bavarian Wikipedia data.\footnote{\url{https://github.com/google-research/bert/blob/master/multilingual.md}}

Since the UDPipe and Stanza pipelines trained on HDT perform markedly worse than their GSD counterparts, we only use GSD for finetuning our models.
We use the regular GSD treebank, as well as a version with character-level noise inspired by~\citet{aepli-sennrich-2022-improving} to see if it improves over vanilla fine-tuning. For character-level noise, we select a certain ratio of the words (`noise level') in each sentence and randomly inject noise into each word by replacing, deleting or inserting a character.
We use the \textit{split word ratio difference} heuristic \citep{blaschke-etal-2023-manipulating} for selecting appropriate noise levels:
For each noise level in \{0, 10, 20, ..., 100\} and each pretrained language model's tokenizer, we compare the proportion of words that the tokenizer splits into multiple subword tokens in the (noised) German and (untouched) Bavarian data, and select the noise level that minimizes the difference between the two ratios.\footnote{We determine the noise levels based on comparisons with just 10\% of the sentences in MaiBaam, in order not to overfit to our test data. However, we observe that the split word ratios are very similar if we use the full dataset.}
For mBERT and {\xlmr}, this means we inject noise into 40\perc{} of the words in a sentence, and 50\perc{} for GBERT.
We train each model on three random seeds and report the mean results in the next section.

\subsection{Results}
\renewcommand{\arraystretch}{1.3}
\begin{table}
\begin{adjustbox}{max width=\columnwidth, center}
\begin{tabular}{@{}llrrrr@{}}
\toprule
 & & \multicolumn{2}{c}{\textbf{POS}}& \multicolumn{2}{c}{\textbf{Dependency}} \\ 
 \cmidrule(lr){3-4} \cmidrule(lr{-1pt}){5-6}
\textbf{Model} & \textbf{Train}  & \multicolumn{1}{c}{\textbf{Acc}} & \multicolumn{1}{c}{\textbf{{\f}}}& \multicolumn{1}{c}{\textbf{LAS}} & \multicolumn{1}{c}{\textbf{UAS}} \\
  \midrule
UDPipe & GSD & \cellcolor[HTML]{57BB8A}{\color[HTML]{FFFFFF} 80.29} & \cellcolor[HTML]{8AD0AE}62.45 & \cellcolor[HTML]{80CCA7}65.79 & \cellcolor[HTML]{59BC8C}{\color[HTML]{FFFFFF} 79.60} \\
UDPipe & HDT & \cellcolor[HTML]{63C092}{\color[HTML]{FFFFFF} 76.36} & \cellcolor[HTML]{93D3B4}59.30 & \cellcolor[HTML]{8CD1AF}61.55 & \cellcolor[HTML]{6AC398}{\color[HTML]{FFFFFF} 73.59} \\
\\[\spacer]
Stanza & GSD & \cellcolor[HTML]{C3E7D5}42.30 & \cellcolor[HTML]{D2EDE0}36.73 & \cellcolor[HTML]{F4FBF7}24.89 & \cellcolor[HTML]{C8E9D9}40.39 \\
Stanza & HDT & \cellcolor[HTML]{CAEADA}39.80 & \cellcolor[HTML]{D4EEE1}36.10 & \cellcolor[HTML]{FFFFFF}20.67 & \cellcolor[HTML]{E8F6EF}29.04 \\
\\[\spacer]
mBERT & GSD\textsubscript{n40} & \cellcolor[HTML]{5CBD8E}{\color[HTML]{FFFFFF} 78.74} & \cellcolor[HTML]{94D4B5}58.74 & \cellcolor[HTML]{9FD8BC}54.96 & \cellcolor[HTML]{7FCBA6}66.38 \\
mBERT & GSD & \cellcolor[HTML]{5FBF90}{\color[HTML]{FFFFFF} 77.47} & \cellcolor[HTML]{99D6B8}57.19 & \cellcolor[HTML]{A6DBC1}52.48 & \cellcolor[HTML]{85CEAA}64.30 \\
GBERT & GSD\textsubscript{n50} & \cellcolor[HTML]{67C296}{\color[HTML]{FFFFFF} 74.68} & \cellcolor[HTML]{99D6B8}57.15 & \cellcolor[HTML]{ABDDC5}50.57 & \cellcolor[HTML]{89D0AD}62.67 \\
GBERT & GSD & \cellcolor[HTML]{94D4B5}58.86 & \cellcolor[HTML]{B5E1CB}47.21 & \cellcolor[HTML]{D3EEE1}36.40 & \cellcolor[HTML]{ABDDC5}50.51 \\
XLM-R & GSD\textsubscript{n40} & \cellcolor[HTML]{6EC49A}{\color[HTML]{FFFFFF} 72.45} & \cellcolor[HTML]{9ED8BB}55.33 & \cellcolor[HTML]{B0DFC8}48.81 & \cellcolor[HTML]{8DD1B0}61.25 \\
XLM-R & GSD & \cellcolor[HTML]{9FD8BC}55.00 & \cellcolor[HTML]{BEE5D2}44.03 & \cellcolor[HTML]{E1F3EA}31.42 & \cellcolor[HTML]{BFE6D3}43.42\\
\bottomrule
\end{tabular}
\end{adjustbox}
    \caption{\textbf{Prediction scores on \projname, in~\%.} The {\f} scores are macro-averaged. Except for the UDPipe and Stanza results, all values are averaged over three runs. The subscript additions \textsubscript{n40} and  \textsubscript{n50} refer to noise levels of 40 and 50\perc{}, respectively (see \S\ref{sec:models} for details).}
    \label{tab:results}
\end{table}

\begin{figure*}
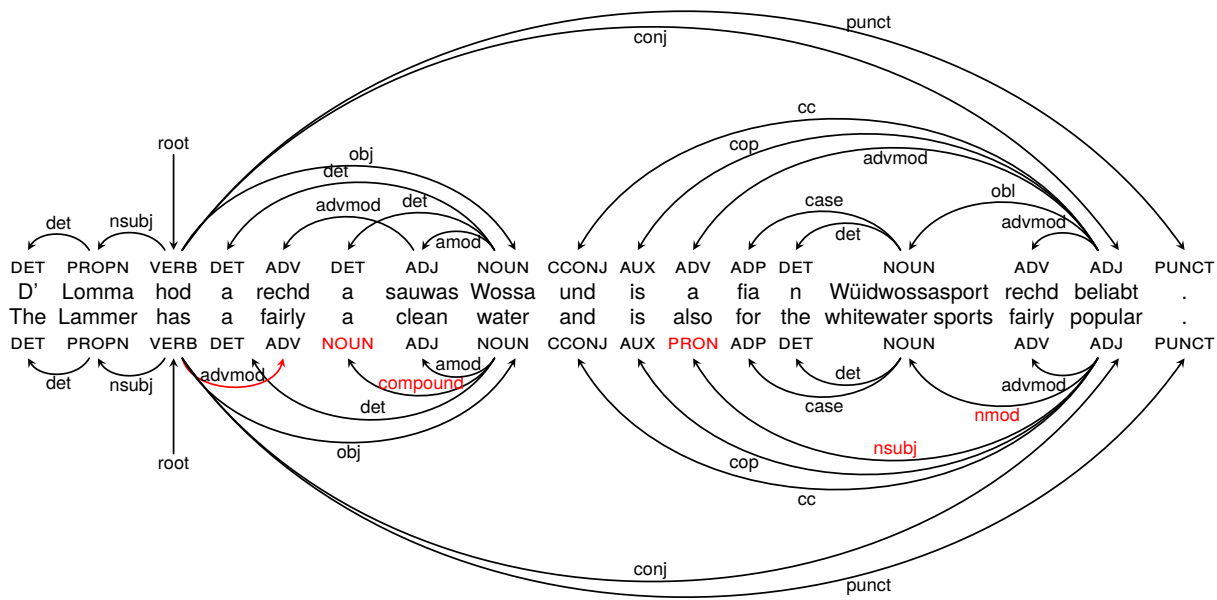

\vspace{-2\baselineskip}
\noindent
\begin{adjustbox}{max width=\textwidth, center}
\hspace{-1em}
\begin{dependency}
[theme=simple,
label style={font=\large},
edge style={thick},
arc angle=60,
]
\begin{deptext}
\pos{det} \& \pos{propn} \& \pos{verb} \& \pos{det} \& \pos{adv} \& \pos{det} \& \pos{adj} \& \pos{noun} \& \pos{cconj} \& \pos{aux} \& \pos{adv} \& \pos{adp} \& \pos{det} \& \pos{noun} \& \pos{adv} \& \pos{adj} \& \pos{punct}\\
D'\& Lomma \& hod \& a \& rechd \& a \& sauwas \& Wossa \& und \& is \& a \& fia\& n \& Wüidwossasport \& rechd \& beliabt\&. \\
The\& Lammer\& has \& a \& fairly\& a\& clean\& water\& and\& is\& also\& for\& the\& whitewater sports\& fairly\& popular\& .\\
\pos{det} \& \pos{propn} \& \pos{verb} \& \pos{det} \& \pos{adv} \& \wrongpos{noun} \& \pos{adj} \& \pos{noun} \& \pos{cconj} \& \pos{aux} \& \wrongpos{pron} \& \pos{adp} \& \pos{det} \& \pos{noun} \& \pos{adv} \& \pos{adj} \& \pos{punct}\\
\end{deptext}
\depedge{2}{1}{det}
\depedge{3}{2}{nsubj}
\deproot{3}{root}
\depedge[left=4mm]{8}{4}{det}
\depedge{7}{5}{advmod}
\depedge[left=1mm]{8}{6}{det}
\depedge[label style={below}]{8}{7}{amod}
\depedge[right=2mm, edge end x offset=2mm]{3}{8}{obj}
\depedge[left=5mm]{16}{9}{cc}
\depedge[label style={below}, left=20mm]{16}{10}{cop}
\depedge[label style={below}]{16}{11}{advmod}
\depedge{14}{12}{case}
\depedge[label style={below}]{14}{13}{det}
\depedge{16}{14}{obl}
\depedge[left=5mm]{16}{15}{advmod}
\depedge[arc angle=55, label style={below}, edge end x offset=2mm]{3}{16}{conj}
\depedge[right=30mm, arc angle=55, label style={below}]{3}{17}{punct}
\depedge[edge below, label style={below}]{2}{1}{det}
\depedge[edge below, label style={below}]{3}{2}{nsubj}
\deproot[edge below, label style={below}]{3}{root}
\depedge[edge below, edge end x offset=4mm]{8}{4}{det}
\depedge[edge style={red}, edge below, arc angle=75]{3}{5}{advmod}
\depedge[label style={text=red}, edge below]{8}{6}{compound}
\depedge[edge below]{8}{7}{amod}
\depedge[edge below, label style={below}, edge end x offset=2mm]{3}{8}{obj}
\depedge[edge below, label style={below}, left=5mm]{16}{9}{cc}
\depedge[edge below, left=20mm]{16}{10}{cop}
\depedge[label style={text=red}, edge below]{16}{11}{nsubj}
\depedge[edge below, label style={below}]{14}{12}{case}
\depedge[edge below]{14}{13}{det}
\depedge[left=1mm, label style={text=red, below}, edge below]{16}{14}{nmod}
\depedge[left=5mm, edge below, label style={below}]{16}{15}{advmod}
\depedge[edge below, arc angle=55, edge end x offset=2mm]{3}{16}{conj}
\depedge[right=30mm, edge below, arc angle=55]{3}{17}{punct}
\end{dependency}
\end{adjustbox}
\vspace{-3\baselineskip}
\caption{\textbf{Gold-standard (top) and predicted (bottom) annotations.}
Predictions are produced by the UDPipe model trained on GSD, the best system in our evaluation.
Wrong predictions are in red.\\
\translationlb{`The Lammer (river) has fairly clean water and is also pretty popular for whitewater sports.'}{(Wiki \textit{Låmma} `Lammer')}
}
\label{fig:gold-pred}
\end{figure*}

Table~\ref{tab:results} shows the different models' dependency parsing and POS tagging scores.
For dependency parsing, we use the unlabelled and labeled attachment scores (UAS and LAS, respectively),\footnote{For LAS, we ignore relation subtypes, as in UD's official evaluation script: \href{https://github.com/UniversalDependencies/tools/blob/master/eval.py}{\texttt{github.com/\allowbreak{}Universal\allowbreak{}Dependencies/\allowbreak{}tools/\allowbreak{}blob/\allowbreak{}master/\allowbreak{}eval.py}}.} and for POS tagging we consider both accuracy and macro-averaged {\f} scores.
Figure~\ref{fig:gold-pred} shows a parse produced by the best system. %
While most of its predictions are correct, they contain several wrong POS tags and dependencies (arcs and labels). Half of the prediction errors in this sentence affect the phrase with the doubled determiner, which is not licensed by Standard German grammar (\S\ref{sec:np}; \textit{a rechd a sauwas Wossa} `lit.: a fairly a clean water').

For both UDPipe and Stanza, we observe that the versions trained on the GSD treebank outperform those trained on the larger HDT.
The UDPipe models (regardless of training data) attain the highest scores for all four metrics (except for the HDT version's POS tagging accuracy).
The single best model is the UDPipe version trained on GSD, reaching LAS and UAS scores of 65.79\perc{} and 79.60\perc{}, and, for POS tagging, an accuracy 80.29\perc{} and {\f} score of 62.45\perc.
Conversely, the Stanza models do not generalize well to the Bavarian data and achieve the worst scores of all models across all metrics.\footnote{When evaluating the GSD and HDT Stanza models on the test splits of their respective training sets, all scores are very high and similar to those of UDPipe.}

Focusing on the models we trained on the vanilla GSD data, we observe that mBERT outperforms the other two language models.
Presumably, it benefits from the overlap between some of our corpus data and its (unlabelled) pretraining data.
Including many languages \textit{other} than Bavarian in the pretraining data does not appear to be nearly as advantageous, as the German model GBERT produces better results than the multilingual {\xlmr}.

Injecting noise into the training data consistently improves our models' performance.
The improvements are especially large for the worse-performing {\xlmr} and GBERT. For {\xlmr}, the scores are between 26 and 55\perc{} (11.31--17.83 percentage points; pp) higher with noise than without, and GBERT sees improvements of 21 to 39\perc{} (9.94--15.82~pp).
Conversely, mBERT's scores improve only by between 2~and 5\perc{} (1.27--2.48~pp).

\subsection{Discussion}

The POS tagging scores of our best re-trained model (mBERT with noised data) are competitive with those of the UDPipe models.
We hypothesize that UDPipe's character embeddings are of advantage when processing the orthographically very variable Bavarian input.
Stanza, on the other hand, uses static German embeddings for entire words -- a mismatch for the Bavarian input.
The subword tokens used by mBERT, GBERT and {\xlmr} provide an intermediary input representation, and the versions finetuned on the noised data are better geared towards processing short subwords.
However, the dependency parsing results of our models clearly lag behind UDPipe.
We assume that this is owed to UDPipe's processing steps for properly constructing directed spanning dependency graphs. 

\section{Conclusion}

We present \projname, the first Bavarian treebank, which we manually annotated with POS tags and syntactic dependencies in UD.
It comprises a range of dialects and genres. 
We share \projname{} as a resource for analyzing and modelling Bavarian data and, more broadly, non-standard language data.

We also conduct transfer learning experiments with models trained on German data to provide parsing and POS tagging baselines for \projname.
Even the best model has ample room for improvement. This shows that processing Bavarian data is not as simple as merely using zero-shot transfer from German.

\section*{Acknowledgements}

We thank Miriam Winkler and Marie Kolm for lending us their native speaker expertise.
We also thank the anonymous reviewers for their feedback.

This research is supported by the ERC Consolidator Grant DIALECT 101043235.
We also gratefully
acknowledge partial funding by the European Research Council (ERC
\#740516).

\section{Bibliographical References}\label{sec:reference}
\bibliographystyle{lrec-coling2024-natbib}
\bibliography{bibliography}

\section{Language Resource References}\label{lr:ref}
\bibliographystylelanguageresource{lrec-coling2024-natbib}
\bibliographylanguageresource{languageresource, ud}

\newpage %
\appendix
\section{Data Statement}
\label{sec:data-statement}

\paragraph{Header}

\begin{itemize}
\item \textit{Dataset title:} MaiBaam
\item \textit{Dataset curator(s):} Verena Blaschke, Barbara Kova\v{c}i\'{c}, 
Siyao Peng, 
Barbara Plank
\item \textit{Dataset version:} 1.0 (UD version~2.14)
\item \textit{Dataset citation:} MaiBaam should be cited by citing this article.
\item \textit{Data statement authors:} Verena Blaschke, Barbara Kova\v{c}i\'{c}, 
Siyao Peng, 
Hinrich Schütze, 
Barbara Plank
\item \textit{Data statement version:} 1.0
\item \textit{Data statement citation and DOI:} To cite this data statement, please cite this publication.
\item \textit{Links to versions of this data statement in other languages:} ---
\end{itemize}

\paragraph{Executive Summary}

{\projname} is a manually annotated dependency treebank for Bavarian.
It contains 15k tokens and is annotated with part-of-speech tags and syntactic dependencies according to Universal Dependency guidelines.\footnote{\url{https://universaldependencies.org/bar/}}
MaiBaam encompasses diverse text genres (wiki articles and discussions, grammar examples, fiction, and commands for virtual assistants) and dialects from the North, Central and South Bavarian areas as well as the dialectal transition areas in between.
In addition, it provides sentence-level genre and dialect metadata.

\paragraph{Curation Rationale}

The purpose of {\projname} is to allow research on computational methods for processing non-standardized language data, including the evaluation of cross-lingual transfer setups (given the large amount of German and other data also annotated according to UD guidelines).
Furthermore, it allows researching syntactic structures of Bavarian (on their own, and in contrast to the other Germanic languages covered by UD).
Our goal is to represent as many Bavarian dialects and as many text genres as possible given the availability and licensing of such data.
Each data instance is a sentence, annotated with part-of-speech tags and syntactic dependencies.

\paragraph{Documentation for Source Datasets}

MaiBaam contains sentences from the South Tyrolean \citep{winkler2024slot} and Central Bavarian \citep{winkler2024slot} versions of xSID (\href{https://bitbucket.org/robvanderg/xsid/src/master/LICENSE}{CC BY-SA 4.0 International}) as well as additional data from \citet{winkler2024slot} (to be released soon, likely under the same license), Tatoeba (\href{https://creativecommons.org/licenses/by/2.0/fr/}{CC-BY 2.0 FR}), Wikipedia (\href{https://creativecommons.org/licenses/by-sa/4.0/}{CC BY-SA 4.0 International}), and UD's Cairo CICLing Corpus (no license).

\paragraph{Language Varieties}

MaiBaam contains Bavarian data (ISO~639-3: \href{https://iso639-3.sil.org/code/bar}{bar}, Glottocode: \href{https://glottolog.org/resource/languoid/id/bava1246}{bava1246}, BCP-47: bar-DE, bar-AT, bar-IT) from the North, Central and South Bavarian areas as well as the transition areas between North and Central Bavarian and Central and South Bavarian.
Details are in Table~\ref{tab:dialects}.

\paragraph{Speaker Demographic}

We do not have detailed demographic information on the speakers whose data we include in \projname, with the exception of differently granular geographic and/or dialectal information (this applies to 80\perc{} of the utterances).
About 17\perc{} of the tokens (13\perc{} of the sentences) belong to dialects spoken in larger cities (Munich, Vienna, Salzburg).
Details on the geographic and dialectal distribution can be found in Table~\ref{tab:dialects}.

\paragraph{Annotator Demographic}

All involved in providing Bavarian translations and helping us with questions about Bavarian words/sentences or linguistic structures are native speakers of Bavarian (two Central Bavarian speakers from Bavaria, one South Bavarian speaker from South Tyrol).

The annotator is a native speaker of German and a (non-Bavarian) Upper German dialect.
The annotation guidelines were created and refined by a native speaker of German familiar with German dialectology, a learner of German with experience in annotating UD treebanks, and the annotator.
They were also reviewed by a native speaker of a Bavarian dialect.

Everybody involved in this project, except for one of the Bavarian informants, has a background in (computational) linguistics.
Additional details are in Section~\ref{sec:annotation-procedure}.

\paragraph{Speech Situation and Text Characteristics}

\begin{itemize}
\item \textit{Time of linguistic activity:}
Wikipedia articles and discussions: unknown between 2006--2024; Tatoeba sentences: 2013--2022; Fairy tales: 2018; xSID: translated in 2021 (South Tyrolean) and 2023 (data from Bavaria); other virtual assistant data: 2023; CICLing: translated in 2023
\item \textit{Date of data collection:} 2023--2024
\item \textit{Modality:} Written
\item \textit{Synchronous vs.\ asynchronous interaction:} Asynchronous
\item \textit{Scripted/edited vs.\ spontaneous:} Presumably edited in most cases. The CICLing and xSID sentences are translations, as are many of the Tatoeba sentences. Some of the sentences from Wikipedia articles or from fairy tales might be translations.
\item \textit{Speakers' intended audience:} The sentences in xSID are queries for a hypothetical digital assistant. The linguistic example sentences from CICLing and Wikipedia articles are for people interested in linguistics. All other data are for an audience of internet users who are interested in reading Bavarian content, be they themselves speakers of Bavarian or not.
\item \textit{Genre:} See Section~\ref{sec:corpus-data} and Table~\ref{tab:genres}.
\item \textit{Topic:} Various.
The wiki articles include locations, traditions/customs, food and entertainment/media, among other topics.
\item \textit{Non-linguistic context:} ---
\end{itemize}

\paragraph{Preprocessing and Data Formatting}

We manually replace usernames mentioned in Wikipedia discussions with \textsc{username}.
We ignore the original text formatting choices (italics, boldface, typeface).
We do not include the raw, unannotated data in the dataset.
The dataset adheres to \href{https://universaldependencies.org/format.html}{\makebox{CoNLL-U}} formatting.

\paragraph{Capture Quality}

No known issues.

\paragraph{Limitations}

We cannot verify that all sentences were written by competent speakers of Bavarian.

\paragraph{Metadata}

\begin{itemize}
\item \textit{License:} \href{https://bitbucket.org/robvanderg/xsid/src/master/LICENSE}{CC BY-SA 4.0 International}
\item \textit{Annotation guidelines:} \citet{maibaam-guidelines}
\item \textit{Annotation process:} See Section~\ref{sec:annotation-procedure}.
\item \textit{Dataset quality metrics:} --- %
\item \textit{Errata:} None so far. Please report errors by contacting the authors or opening an issue at \repoissues.
\end{itemize}

\paragraph{Disclosures and Ethical Review}

We only collected and annotated data that were shared under licences that explicitly permit adapting and re-sharing the data.
Everyone involved in annotating and translating data and everyone we consulted with questions about Bavarian was hired and compensated according to local standards.
An institutional
ethics review process was not accessible at the time of dataset creation.

There are no conflicts of interest.
This research is supported by the ERC Consolidator Grant DIALECT 101043235.
We also gratefully
acknowledge partial funding by the European Research Council (ERC
\#740516).

\paragraph{Other}

---

\paragraph{Glossary}

---

\paragraph*{About this document}

A data statement is a characterization of a dataset that provides context to allow developers and users to better understand how experimental results might generalize, how software might be appropriately deployed, and what biases might be reflected in systems built on the software.

This data statement was written based on the template for the Data Statements Version~2 Schema. The template was prepared by Angelina McMillan-Major, Emily M. Bender, and Batya Friedman, and can be found at \href{http://techpolicylab.uw.edu/data-statements}{\texttt{http://\allowbreak{}tech\allowbreak{}policy\allowbreak{}lab.\allowbreak{}uw.\allowbreak{}edu/\allowbreak{}data-\allowbreak{}statements}}.

\section{POS Tag Distributions}
\label{sec:label-distributions}
Table~\ref{tab:pos-tags} provides an overview of the part-of-speech tags in \projname, and Table~\ref{tab:deprels} shows the distribution of dependency relations.

\begin{table*}[]
\begin{tabular}{@{}ll@{\hspace{-1.5em}}rrrl@{}}
\toprule
\multicolumn{2}{l}{\textbf{Part-of-speech tag}} & \textbf{\# Tokens} & \% & \textbf{TTR} & \textbf{Top 5 most frequent tokens}\\\midrule
\pos{NOUN} & Noun & 2\,269 & 15.1 & 0.65 & 
\textit{Wecka} `alarm', \textit{Kafää} `coffee', \textit{Mingara}\\
&&&&&`Munich citizen', \textit{Leid} `people', \textit{See} `lake'\\
\pos{PUNCT} & Punctuation & 2\,105 & 14.0 & 0.01 & . , '' ? ( \\
\pos{DET} & Determiner & 1\,946 & 13.0 & 0.11 & \textit{m, da, a, de, an} \\
\pos{VERB} & Verb & 1\,458 & 9.7 & 0.65 & \textit{gibt} `[there] is', \textit{hod} `has', 
\textit{Erinner} `remind', \\
&&&&& \textit{sogt} `says', \textit{gsogt} `said'\\
\pos{ADP} & Adposition & 1\,417 & 9.4 & 0.14 & \textit{in, vo, i, auf, mit}\\
\pos{ADV} & Adverb & 1\,206 & 8.0 & 0.39 & \textit{aa} `also', \textit{so} `so', \textit{a} `also', \textit{no} `still', \textit{do} `there'\\
\pos{PRON} & Pronoun & 1\,127 & 7.5 & 0.11 & \textit{ma} `we, me', \textit{i} `I', \textit{s} `it, she/her, you.\textsc{pl}',\\
&&&&& \textit{I} `I', \textit{des} `this/that'\\
\pos{AUX} & Auxiliary & 926 & 6.2 & 0.21 & \textit{is} `is', \textit{hod} `has', \textit{san} `[we/they] are',\\
&&&&& \textit{hob} `[I] have', \textit{hom} `[we] have'\\
\pos{ADJ} & Adjective & 799 & 5.3 & 0.83 & \textit{neie} `new', \textit{guat} `good', \textit{Soizburga}\\
&&&&& `of Salzburg', \textit{guad} `good', \textit{gaunzn} `entire', \\
\pos{PROPN} & Proper noun & 550 & 3.7 & 0.66 & \textit{Minga} `Munich', \textit{Thomas, Tom, Gretl, Hansl}\\
\pos{CCONJ} & Coordinating conjunction & 380 & 2.5 & 0.08 & \textit{und} `and', \textit{oda} `or', \textit{owa} `but', \textit{oder} `or', \\
&&&&&\textit{Und} `and'\\
\pos{SCONJ} & Subordinating conjunction& 341 & 2.3 & 0.23 & \textit{dass} `that', \textit{das} `that', \textit{wia} `than', \textit{ wej} `\textsc{rel}', \\
&&&&& \textit{wos} `\textsc{rel}'\\
\pos{NUM}& Numeral & 240 & 1.6 & 0.58 &  \textit{5,} \textit{zwoa} `two', \textit{4, 6} \textit{drei} `three'\\
\pos{PART} & Particle& 165 & 1.1 & 0.13 & \textit{ned} `not', \textit{net} `not', \textit{zu} `\textsc{inf}', \textit{niat} `not', \\
&&&&&\textit{nim} `not [anymore]'\\
\pos{X}& Other & 64 & 0.4 & 0.89 & \textit{e, d,} \textit{Schuhplattler*} `[dance]', \textit{München*} \\
&&&&& `Munich', \textit{m\i\ng(:)\textturna}\\
\pos{INTJ}& Interjection & 23 & 0.2 & 0.70 & \textit{Gö, gö, Ja} `yes',  \textit{Bfiade} `bye', \textit{Seavas} `hi/bye'\\
\pos{SYM}& Symbol & 7 & 0.0 & 1.00 & *, †, \%, <, :-)\\ \bottomrule
\end{tabular}
\caption{\textbf{POS tag statistics.} For each POS tag, we provide the absolute and relative (\%, in percent) number of tokens, the type-token ratio (TTR) and the most frequent tokens.
The asterisk* denotes German words that are clearly presented as non-Bavarian material in a given sentence.}
\label{tab:pos-tags}
\end{table*}

\begin{table*}
\centering
\begin{tabular}{@{}l@{}rr@{\hspace{5em}}l@{}rr@{\hspace{5em}}l@{}rr@{}}
\toprule
\textbf{Relation} & \multicolumn{1}{l}{\textbf{Abs.}} & \multicolumn{1}{l}{\textbf{~~\%}} & \textbf{Relation} & \multicolumn{1}{l}{\textbf{Abs.}} & \multicolumn{1}{l}{\textbf{~~\%}} & \textbf{Relation} & \multicolumn{1}{l}{\textbf{Abs.}} & \multicolumn{1}{l}{\textbf{~~\%}} \\ \midrule
punct & 2105 & 14.0 & appos & 176 & 1.2 & nsubj:pass & 66 & 0.4 \\
det & 1746 & 11.6 & advcl & 174 & 1.2 & fixed & 41 & 0.3 \\
advmod & 1403 & 9.3 & nummod & 140 & 0.9 & acl & 31 & 0.2 \\
case & 1329 & 8.8 & flat & 126 & 0.8 & orphan & 30 & 0.2 \\
nsubj & 1128 & 7.5 & expl & 110 & 0.7 & discourse & 23 & 0.2 \\
root & 1070 & 7.1 & expl:pv & 107 & 0.7 & advcl:relcl & 21 & 0.1 \\
obl & 798 & 5.3 & obl:arg & 105 & 0.7 & vocative & 16 & 0.1 \\
obj & 670 & 4.5 & compound:prt & 103 & 0.7 & compound & 15 & 0.1 \\
aux & 575 & 3.8 & ccomp & 99 & 0.7 & obl:agent & 11 & 0.1 \\
amod & 468 & 3.1 & det:poss & 90 & 0.6 & goeswith & 8 & 0.1 \\
conj & 467 & 3.1 & acl:relcl & 88 & 0.6 & csubj & 7 & 0.0 \\
nmod & 446 & 3.0 & parataxis & 84 & 0.6 & dislocated & 7 & 0.0 \\
cc & 377 & 2.5 & aux:pass & 81 & 0.5 & reparandum & 3 & 0.0 \\
mark & 342 & 2.3 & xcomp & 70 & 0.5 & dep & 2 & 0.0 \\
cop & 265 & 1.8 &  & \\ \bottomrule
\end{tabular}
\caption{\textbf{Dependency relation statistics.} For each dependency relation, we provide the absolute (\#) and relative (\%, in percent) number of occurrences.}
\label{tab:deprels}
\end{table*}

\end{document}